\documentclass[runningheads]{llncs}

\usepackage{graphicx}
\usepackage{tikz}
\usepackage{comment}
\usepackage{amsmath,amssymb} 
\usepackage{color}
\usepackage{booktabs} 
\usepackage{multirow}
\usepackage{wrapfig}

\usepackage{hyperref}

\begin{document}
\pagestyle{headings}
\mainmatter
\def\ECCVSubNumber{18}  

\title{Can Everybody Sign Now? \\
Exploring Sign Language Video Generation\\from 2D Poses} 

\titlerunning{Can Everybody Sign Now?}
\author{Lucas Ventura\inst{1} \and
Amanda Duarte\inst{1,2} \and
Xavier Giró-i-Nieto\inst{1,2}}

%
\authorrunning{Ventura et al.}
%
\institute{Universitat Polit\`{e}cnica de Catalunya, Barcelona, Catalonia/Spain
\and
Barcelona Supercomputing Center, Spain
\email{lucas.ventura.ripol@estudiantat.upc.edu \{amanda.duarte,xavier.giro\}@upc.edu}}
\maketitle

\makeatletter{\renewcommand*{\@makefnmark}{}
\footnotetext{Work presented as an extended abstract at the Sign Language Recognition, Translation \& Production (SLRTP) workshop (https://slrtp.com/).}\makeatother}
\section{Introduction}

Sign Language is the primary means of communication of the Deaf community but barely known by the rest of the population.
This situation creates difficulties in conversations between sign and non-sign language speakers, which are normally addressed with textual transcriptions of the spoken language, or the sign-speakers developing lip-reading and oral communication skills.

The communication barrier between sign and non-sign language speakers may be reduced in the coming years thanks to the recent advances in neural machine translation and computer vision.
Recent works~\cite{saunders2020progressive,Text2Sign,WordsAreOurGlosses} are making steps towards sign language translation by automatically generating detailed human pose skeletons from spoken language. Skeletons are represented by 2D/3D coordinates of human joints also known as \textit{keypoints}; given a set of estimated keypoints, one can visualize them as a wired skeleton connecting the modeled joints (see the middle row of Figure~\ref{fig:ESN_merged}). Although such visualizations are theoretically useful for understanding sign language, no studies have been made so far on whether they are indeed understood by deaf people.
 
In this work, we study \textit{if and how well members of the Deaf community understand automatically generated sign language videos}. Apart from skeleton visualizations, we go one step further and generate realistic videos using the state of the art human motion transfer method Everybody Dance Now (EDN)~\cite{EDN}. 
We run a study with four native sign language speakers and record their understanding of both skeleton visualizations and generated signing videos by asking three different types of feedback: a global classification of the video in terms of topic, a translation into American English, and a final subjective rating about how understandable the videos were. We further quantitatively study the quality of the videos generated using the EDN method via the percentage of keypoints one can re-estimate on the generated frames.

For signing videos and keypoints, we utilize a subset of the recent How2Sign~\cite{How2Sign} dataset, a large dataset of American Sign Language (ASL) signing videos.
Our main results indicate that a) the generated videos were generally preferred over the skeleton visualizations and that b) the current state of the art in image generation is not good enough for sign language translation out-of-the-box. Specifically, we show that the model struggles with generating the hands, which play a central role in sign language understanding.

\section{Methodology and Results}

\begin{figure*}[t]
    \centering
    \vspace{-6mm}
    \caption{Sample of the source video (top row) used to automatically extract 2D keypoints with OpenPose~\cite{openpose} (middle row) and generate frames for a target identity (bottom row). A sample of the generated video can be seen at: \url{https://youtu.be/4ve1sGzWl2g}.}
    \label{fig:ESN_merged}
    \includegraphics[width=.9\textwidth]{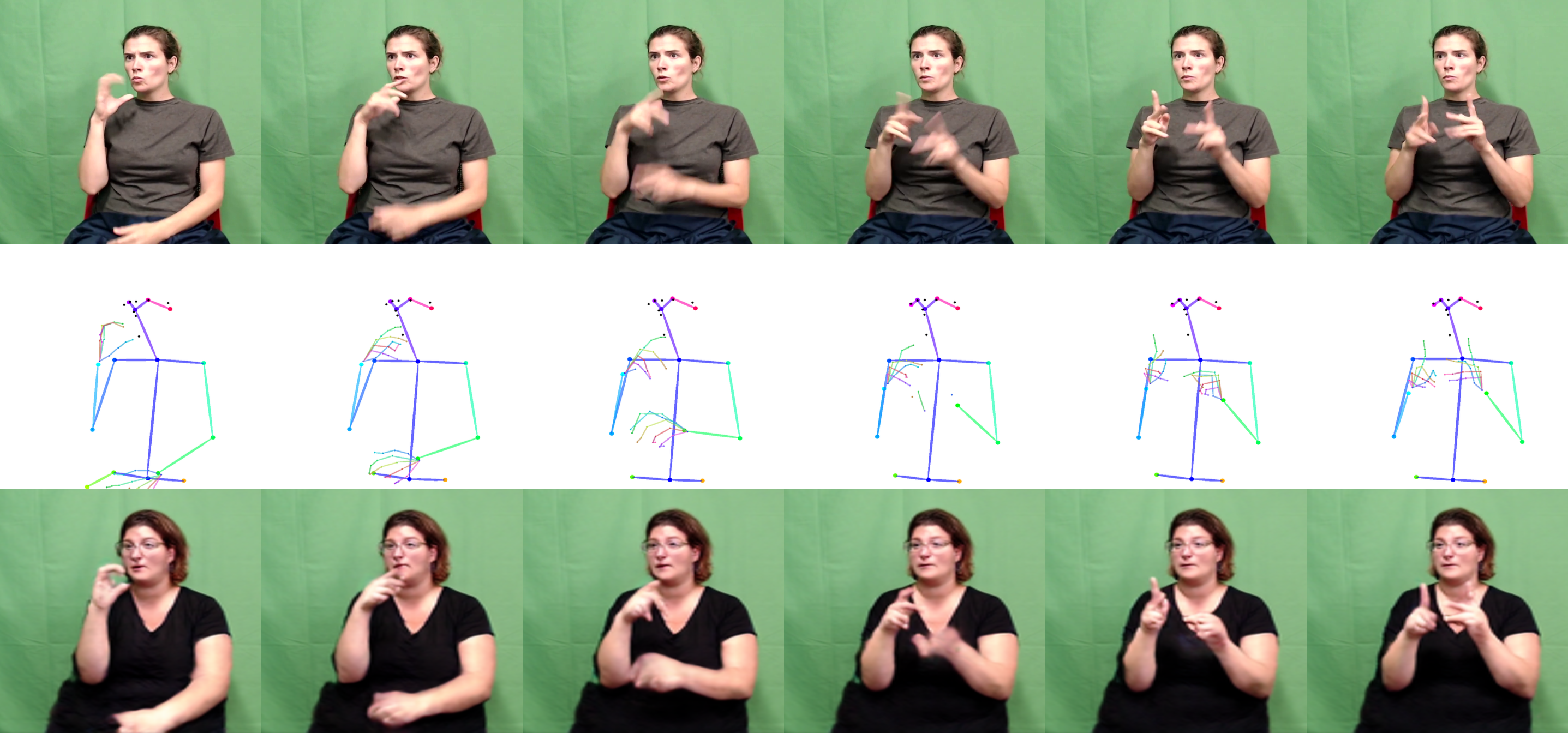}
\end{figure*}

\noindent\textbf{Generating realistic signing videos.}~To generate an animated video of a signer given a set of keypoints, we use the Everybody Dance Now (EDN)~\cite{EDN} approach. It is worth noting that this approach models facial landmarks separately, something highly desirable in our case as they are one of the critical features for sign language understanding. The input and output is shown in Figure \ref{fig:ESN_merged}. We use OpenPose~\cite{openpose} to extract keypoints (middle row) from the source video (top row); the keypoints are then used to condition a Generative Adversarial Network (GAN) that generates each video frame, using the model from~\cite{wang2017highresolution}.

Our model was trained on a subset of the How2Sign dataset~\cite{How2Sign} that contains videos from \textit{two} professional American Sign Language interpreters. Specifically, keypoints extracted from videos of the first signer (top row in Figure~\ref{fig:ESN_merged}) were used to learn the model that generates realistic videos of the second signer (bottom row). The training dataset consists of more than 28 hours of sign language translations from instructional videos.

\noindent\textbf{Quantitative Results.} An approximate but automatic way of measuring the visual quality of the generated videos is by measuring the number of keypoints that can be reliably detected by OpenPose in the source and generated videos. We focus only on the 125 upper body keypoints which are visible in the How2Sign videos and discard those from the legs. We use two metrics: a) the Percentage of Detected Keypoints (PDK), which corresponds to the fraction of keypoints from the source frame which were detected in the synthesized frame and b) the Percentage of Correct Keypoints (PCK)~\cite{PCK}, which labels each detected keypoint as ``correct'' if the distance to the keypoint in the original image is less than 20\% of the torso diameter.
 
In Table \ref{table:table_quantative_results} we present these metrics for different OpenPose confidence thresholds. We report results for all keypoints, as well as when restricting the evaluation only on the hand keypoints, as this is a very important part of sign language understanding. We see that although in general the repeatability of keypoints is high, when focusing on the hands, the model fails to generate reliable keypoints, something that may severely hinder sign language understanding.

\begin{table}
    \begin{center}
    \caption{
    Percentage of Detected Keypoints (PDK) and Percentage of Correct Keypoints (PCK) for all keypoints (125 for upper body, hands and face) and just for the hands (21 for each hand), when thresholding at different detection confidence scores.}
    \label{table:table_quantative_results}
     \resizebox{.8\columnwidth}{!}{
    \setlength{\tabcolsep}{8pt}
        \begin{tabular}{lccccccc}
        \hline\noalign{\smallskip}
        & \multicolumn{3}{c|}{PDK} & \multicolumn{3}{c}{PCK} \\
        min. detection confidence
        & 0 & 0.2 & \multicolumn{1}{c|}{0.5} & 0 & 0.2 & 0.5 \\
        \noalign{\smallskip}
        \hline
        \noalign{\smallskip}
        All keypoints  &  0.99   &  0.88  &  0.87  &   0.90  &  0.94   & 0.96  \\
        Hands  &  0.99  &  0.38  &  0.17  &   0.18  &  0.23  & 0.26 \\
        \hline
        \end{tabular}
     }
    \end{center}
\end{table}


\noindent\textbf{How understandable are the generated signing videos?} 
We evaluate the degree of understanding for both skeleton visualizations and generated videos by showing 3-minute-long videos to four native ASL speakers. Two watched the generated videos, while the other two watched the animated 2D skeletons visualizations. During the evaluation, each subject was asked to: a) classify six videos between ten categories of instructional videos; b) answer the question \textit{``How well could you understand the video?''} on the five-level scale (Bad, Poor, Fair, Good, Excellent); c) watch two trimmed clips from the previously watched video, and translate them into American English. Results averaged over all subjects are presented in Table~\ref{table:table_study_results}. We report accuracy for the classification task, the Mean Opinion Score (MOS) for the five-scale question answers and BLEU~\cite{papineni2002bleu} scores for the American English translations. Qualitative results are shown in Table~\ref{table:table_qualitative_results}.

Overall, results show a preference towards the generated videos rather than the skeleton ones as the former result to higher scores across all metrics. In terms of general understanding of the topic, it seems that both visualizations did relatively well; we see that subjects were able to mostly classify the videos correctly. When it comes to finer grained understanding, however, as measured via the English translations, we see from both Tables~\ref{table:table_study_results} and \ref{table:table_qualitative_results} that the translation task cannot be solved neither with the skeletons nor with the generated video.

\begin{table}
    \begin{center}
    \caption{\label{table:table_study_results}Comparison between skeletons and generated videos in terms of classification (accuracy), mean opinion score (MOS) and translation (BLEU)~\cite{papineni2002bleu}.}
        \begin{tabular}{lcccccc}
        \toprule
        & \; Accuracy \; & \multicolumn{1}{|c}{} \; MOS \; & \multicolumn{1}{|c}{} \; BLEU-1 & \, BLEU-2 & \, BLEU-3 & \, BLEU-4 \\
        \midrule
        Skeleton visualization & 83.3 \% & 2.50 &  10.90 & 3.02 & 1.87 & 1.25   \\
        Generated video & \textbf{91.6 \%} & \textbf{2.58} &  \textbf{12.38} & \textbf{6.71} & \textbf{3.32} & \textbf{1.89}  \\ 
        \bottomrule
        \end{tabular}
    \end{center}
\end{table}
\begin{table}
    \caption{Groundtruth and translations for two clips from a ``Food and Drinks'' class. All subjects were able to correctly identify the class. Two subjects watched the skeleton visualizations, while two different subjects watched the videos generated by EDN.}
    \label{table:table_qualitative_results}
    \resizebox{\columnwidth}{!}{
    \setlength{\tabcolsep}{15pt}
    \begin{tabular}{ll}
        \toprule
        Groundtruth     & \textbf{I'm not going to use a lot, I'm going to use very very little.}   \\ \cline{2-2} 
        \multirow{2}{*}{Skeleton} & That is not too much                                            \\  
        & don't use much, use a little bit                                                          \\ \cline{2-2}
        \multirow{2}{*}{EDN} & Don't use a lot, use a little                                        \\
         & dont use lot use little bit                                                              \\ 
         \midrule
        Groundtruth       & \textbf{I'm going to dice a little bit of peppers here.}                \\ \cline{2-2}
        \multirow{2}{*}{Skeleton} & cooking                                                         \\
         & chop yellow peppers                                                                      \\ \cline{2-2}
        \multirow{2}{*}{EDN} & cook with a little pepper                                            \\
         & chop it little bit and sprinkle \\
         \bottomrule     
    \end{tabular}
    }
\end{table}
\vspace{-3mm}

\section{Conclusions}
In this paper we investigate how well members of the Deaf community actually understand keypoint-based visualizations, the output of choice for many recent automatic sign language translation works. Through our study, we show that subjects prefer synthesized realistic videos over skeleton visualizations; we also show that out-of-the-box synthesis methods are not really effective enough and that subjects struggled to understand the signing videos. We partially attribute poor understanding on the bad synthesis of the hands, and believe that future research towards that direction is highly important. 
\noindent\textbf{{Acknowledgments.}}
This work was funded by project TEC2016-75976-R of the Spanish Ministerio de Economía y Competitividad and the European Regional Development Fund. Amanda Duarte has received support from the la Caixa Foundation (ID 100010434) under the fellowship code LCF/BQ/IN18/11660029.
 \vspace{-2mm}

\bibliographystyle{splncs04}
\bibliography{egbib}
\end{document}